\documentclass{article}

\usepackage{amsmath}
\usepackage{graphicx} 
\usepackage{booktabs}
\usepackage{multirow}

 \usepackage[dblblindworkshop, final]{neurips_2026}
\workshoptitle{{SLMs for Agentic Systems, Paris, France, 2026}}

\usepackage[utf8]{inputenc} 
\usepackage[T1]{fontenc}    
\usepackage{hyperref}       
\usepackage{url}            
\usepackage{booktabs}       
\usepackage{amsfonts}       
\usepackage{nicefrac}       
\usepackage{microtype}      
\usepackage{xcolor}         

\title{How Fragile Is On-Device Language Model Safety? Localizing Safety-Critical Parameters for Sparse Fault Analysis}

\author{%
  Muhammad Zeeshan Karamat \\
  Virginia Tech\\
  \texttt{mzeeshan@vt.edu} \\
  \And
  Christiana Chamon \\
  Virginia Tech\\
  \texttt{ccgarcia@vt.edu} \\
}

\begin{document}

\maketitle

\begin{abstract}
As small language models (SLMs) are increasingly deployed on resource-constrained and on-
device platforms, including as components of agentic systems, the integrity of
locally stored model parameters becomes an important safety concern. We investigate whether safety-sensitive behavior in LLaMA-2-7B-Chat is concentrated within a sparse subset of parameters, creating a reduced fault surface for targeted analysis. We study two complementary localization methods: low-rank safety-associated subspace analysis and parameter-level safety--utility importance filtering. Both approaches reveal highly non-uniform safety sensitivity across the network, with the MLP \texttt{down\_proj} consistently emerging as a prominent safety-sensitive component and \texttt{o\_proj} providing a smaller contribution. Using parameter-level localization, modifying only 0.19\% of model weights in \texttt{down\_proj} yields 53\% Basic ASR and 56\% GCG ASR, while tinyBenchmarks accuracy remains at 51.6\% compared with a 52.2\% unmodified baseline. These results motivate targeted fault analysis and selective integrity protection for language models deployed in resource-constrained, on-device, and agentic settings.

\end{abstract}

\section{Introduction}

Language models are increasingly being deployed on edge and on-device
platforms, including as components of resource-constrained agentic systems,
offering benefits such as lower latency, improved privacy, and reduced
dependence on cloud connectivity. However, local deployment also exposes
model parameters to hardware-level reliability and integrity threats.
Unlike prompt-based jailbreaks, faults in stored model weights can directly
alter mechanisms responsible for aligned behavior and may persist across
subsequent model interactions. This motivates a central question:
\textit{can model safety be selectively degraded through sparse parameter
modifications without broadly disrupting model functionality?}

A practical bit-level attack on a billion-parameter model faces a large
search problem: only a small fraction of model parameters may be directly
relevant to safety behavior. We therefore adopt a localization-first
perspective. If safety-sensitive behavior depends disproportionately on a
structured subset of the parameter space, these regions may provide a
substantially reduced candidate space for hardware fault analysis.
Crucially, such localization should distinguish safety-associated parameters
from those that are also important for general model utility, since
indiscriminate corruption may simply degrade overall model quality.

We study this question in LLaMA-2-7B-Chat using two complementary
approaches. First, we identify low-rank safety-associated subspaces by
contrasting representations derived from safety and utility calibration
data. Second, we perform parameter-level localization by identifying
weights that are highly important for safety but comparatively less
important for utility. Across both approaches, the MLP
\texttt{down\_proj} consistently emerges as a prominent safety-sensitive
component, with a smaller contribution from the attention
\texttt{o\_proj}. Using parameter-level localization, modifying only
0.19\% of model weights in \texttt{down\_proj} yields 53\% Basic ASR
and 56\% GCG ASR, while tinyBenchmarks accuracy remains at 51.6\%
compared with a 52.2\% unmodified baseline.

Our contributions are threefold: (1) we analyze safety localization at
both subspace and parameter levels; (2) we show that safety-sensitive
parameter support is strongly concentrated in specific architectural
components, particularly \texttt{down\_proj}; and (3) we demonstrate that
substantial safety degradation can be induced by modifying less than
0.2\% of model weights while largely preserving measured general utility,
motivating a reduced search space for future bit-level fault analysis and
selective integrity protection in on-device language models, including
those used as components of resource-constrained agentic systems.

\subsection{Threat Model}
We consider a white-box adversary targeting a safety-aligned language model
deployed locally on an edge or on-device platform, including models that may serve as components of resource-constrained agentic systems. The adversary has access to the model parameters and may perform offline analysis to identify safety-sensitive regions, but does not retrain or fine-tune the model. The objective is to induce sparse parameter perturbations that
degrade safety-aligned behavior while minimizing degradation of benign
model utility.

This threat is particularly acute for SLMs deployed as components
of agentic systems. Unlike a single-turn chat interaction, an
agent's safety behavior is invoked repeatedly and autonomously across a
session, at each planning step, tool call, and intermediate action,
often without a human in the loop to catch an unsafe output. A
parameter-level fault therefore does not need to be re-triggered per
turn, as a prompt-based jailbreak would: once induced, it persists
silently across every subsequent invocation of the model within the
agent loop, including invocations the user never directly observes
(for example, internal planning or tool-selection steps). This makes the
integrity of locally stored weights a distinct concern from
prompt-level robustness in agentic deployments.

Our fault model does not require the adversary to modify inference-time
prompts. We nevertheless evaluate the perturbed model under both standard
harmful prompts and GCG adversarial suffixes, with the latter serving as a
stress test of whether the induced safety degradation persists under a
stronger adversarial prompting setting.
The current experiments simulate parameter-level interventions rather than
physical hardware faults. We therefore study localization of a reduced
safety-sensitive parameter space as a precursor to fault analysis; mapping
these parameters to a small number of physically realizable bit flips under
specific numerical formats, memory architectures, and fault mechanisms is
left to future work.

\section{Related Work}
\paragraph{Small and on-device language models for agentic systems.}
Recent work has highlighted small language models (SLMs) as efficient building
blocks for agentic systems. \citet{belcak2025small} argue that SLMs are
particularly well suited to specialized and repetitive tasks within agentic
workflows, where reduced computational cost can enable more efficient
deployment. Complementary work on MobileLLM
\citep{liu2024mobilellm} develops sub-billion-parameter models specifically
for mobile deployment and demonstrates their capability on common on-device
tasks, including API calling. Although there is no universally agreed-upon
parameter threshold for SLMs, recent literature commonly includes models
with up to approximately 7B parameters, with LLaMA-2-7B and Mistral-7B
serving as representative examples~\citep{corradini2025slm}. Recent systems
work further demonstrates that models at this scale can be executed locally
on resource-constrained devices, including smartphones~\citep{xue2024powerinfer2}.
These trends motivate examining the integrity of safety mechanisms when
language models are deployed locally as components of resource-constrained
and agentic systems.

\paragraph{Safety localization in language models.}
Recent work has identified localized internal mechanisms associated with
refusal and safety alignment. \citet{arditi2024refusal} showed that refusal
behavior in instruction-tuned language models can be mediated by a
low-dimensional direction in the residual stream. \citet{wei2024brittleness}
further identified sparse safety-relevant regions in parameter space at both
parameter and rank levels while disentangling them from regions important for
general utility. Building on this safety--utility perspective, we investigate
whether localized safety-sensitive regions can provide a reduced candidate
space for hardware-level fault analysis in on-device language models.

\paragraph{Bit-flip and fault attacks on neural networks.}
Bit-flip attacks modify the binary representation of model parameters stored
in memory. \citet{rakin2019bitflip} introduced Progressive Bit Search, showing
that a small number of targeted bit flips can severely degrade quantized
neural networks. More recently, \citet{coalson2024prisonbreak} introduced
PrisonBreak, demonstrating targeted bit-flip attacks against safety-aligned
language models while largely preserving benign functionality.
\citet{zahran2025jailbreaking} further studied fault-induced jailbreaking
across FP16, FP8, INT8, and INT4 representations, showing that numerical
format affects both attack effectiveness and vulnerable components. Unlike
approaches that directly search for attack-effective bits, our work first
localizes parameters that are disproportionately important for safety
relative to utility, with the goal of exposing a compact fault surface for
subsequent bit-level analysis.

\section{Methodology}
\label{sec:method}

Our goal is to identify a small subset of model parameters that contribute
disproportionately to safety-aligned behavior while remaining comparatively
less important for general model utility. We study this using two
complementary approaches: (1) low-rank safety subspace localization, which
identifies safety-associated directions in representation space, and
(2) parameter-level safety--utility localization, which identifies
individual safety-sensitive weights.

We build on the safety--utility localization framework of
\citet{wei2024brittleness} and investigate whether the resulting localized
regions can provide a reduced candidate space for hardware fault analysis
in on-device language models.

\subsection{Calibration Datasets}

We use two calibration datasets representing safety-aligned and
general-purpose behavior. The safety set $\mathcal{D}_{s}$ consists of
harmful instructions from AdvBench~\citep{zou2023universal}, paired with
the model's safety-aligned refusal responses. The utility set
$\mathcal{D}_{u}$ uses Alpaca-Cleaned, a cleaned version of Stanford
Alpaca~\citep{taori2023alpaca}, containing benign instruction--response
pairs with safety-related examples removed. For both localization
approaches, prompt tokens are masked and only response-token activations
are used to characterize safety- and utility-associated behavior.

\subsection{Low-Rank Safety Subspace Localization}

We first characterize safety- and utility-associated representation
subspaces within individual transformer projection modules. For a linear
module with weight matrix
$W\in\mathbb{R}^{d_{\mathrm{out}}\times d_{\mathrm{in}}}$, we collect
response-token input activations $X_s$ and $X_u$ independently from
$\mathcal{D}_{s}$ and $\mathcal{D}_{u}$. We form the corresponding
module outputs
\begin{equation}
A_s = X_s W^\top,
\qquad
A_u = X_u W^\top ,
\end{equation}
and compute truncated singular value decompositions
\begin{equation}
A_s \approx U_s\Sigma_sV_s^\top,
\qquad
A_u \approx U_u\Sigma_uV_u^\top .
\end{equation}

Using the right singular vectors, we construct output-space projectors
\begin{equation}
P_s = V_sV_s^\top,
\qquad
P_u = V_uV_u^\top,
\qquad
P_u^\perp = I-P_u .
\end{equation}
We define a utility-orthogonalized safety-associated weight component as
\begin{equation}
\Delta W = P_u^\perp P_s W,
\label{eq:safety_delta}
\end{equation}
and modify the module according to
\begin{equation}
W' = W-\Delta W.
\end{equation}
This construction suppresses a weight component associated with
safety-aligned behavior while reducing overlap with directions identified
from benign utility data.

To examine whether this intervention can be localized to a smaller
parameter set, we sparsify $\Delta W$ by globally ranking its entries by
absolute magnitude. For a retained fraction $\rho$, we define
\begin{equation}
\Delta W^{(\rho)}_{ij}
=
\Delta W_{ij}
\mathbb{I}\!\left[
|\Delta W_{ij}| \geq \tau_\rho
\right].
\end{equation}
where $\tau_\rho$ is the global threshold corresponding to the desired
fraction. The sparse intervention is then
$W'=W-\Delta W^{(\rho)}$. We evaluate this procedure across all
projection modules and under restrictions to specific architectural
components.

\subsection{Parameter-Level Safety--Utility Localization}

Our second approach directly identifies individual weights that are
important for safety-aligned behavior but comparatively less important
for benign utility. For a projection module with weight matrix
$W\in\mathbb{R}^{d_{\mathrm{out}}\times d_{\mathrm{in}}}$, let
$X_{\mathrm{in}}\in\mathbb{R}^{d_{\mathrm{in}}\times n}$ denote its
response-token input activations over a calibration dataset. Following
an activation-aware importance criterion, we assign each weight
$W_{ij}$ an importance score
\begin{equation}
I(W_{ij})
=
|W_{ij}|
\left\|X_{\mathrm{in},j,:}\right\|_2,
\label{eq:importance_score}
\end{equation}
where $X_{\mathrm{in},j,:}\in\mathbb{R}^{n}$ denotes the activations of
input feature $j$ across the calibration tokens, and
$\left\|X_{\mathrm{in},j,:}\right\|_2$ is its $\ell_2$ norm. Thus, the
importance score combines the magnitude of each weight with the
activation magnitude of its corresponding input feature.

We compute Eq.~\ref{eq:importance_score} independently on
$\mathcal{D}_{s}$ and $\mathcal{D}_{u}$, producing safety scores $I_s$
and utility scores $I_u$, respectively. Let $\mathcal{S}_s$ denote the
set of highest-ranked weights under $I_s$, and let $\mathcal{S}_u$
denote the corresponding utility-important set. We define the
safety-specific parameter set as
\begin{equation}
\mathcal{S}_{\mathrm{spec}}
=
\mathcal{S}_s \setminus \mathcal{S}_u.
\label{eq:set_difference}
\end{equation}
This set difference removes weights that are important for both safety
and general instruction-following behavior, leaving candidates that are
comparatively more specific to safety.

We evaluate the functional effect of these parameters through sparse pruning:
\begin{equation}
w'_{ij} =
\begin{cases}
0, & (i,j)\in\mathcal{S}_{\mathrm{spec}},\\
w_{ij}, & \text{otherwise}.
\end{cases}
\end{equation}
The resulting sparse masks identify individual candidate parameters whose
removal disproportionately affects safety-associated behavior relative to
the utility calibration set.

\subsection{Architectural Localization and Sparse Intervention}

Beyond parameter sparsity, we examine whether safety-sensitive structure is
concentrated within particular transformer components. We consider the
attention projections (\texttt{q\_proj}, \texttt{k\_proj},
\texttt{v\_proj}, and \texttt{o\_proj}) and MLP projections
(\texttt{gate\_proj}, \texttt{up\_proj}, and \texttt{down\_proj})
across transformer layers.

For the low-rank approach, we restrict the projected weight intervention
$\Delta W^{(\rho)}$ to different groups of projection modules. For
parameter-level localization, we similarly restrict pruning of
$\mathcal{S}_{\mathrm{spec}}$ to selected projection types and
combinations of components. This analysis tests whether safety-sensitive
interventions require broad modification of the network or can be
localized to specific architectural regions, thereby reducing the
candidate space for subsequent fault analysis.

\section{Experimental Setup}

\paragraph{Model and data.}
We conduct all experiments on LLaMA-2-7B-Chat
\citep{touvron2023llama2}, which falls within the commonly used
parameter range for SLMs (Section~2) and provides a well-characterized,
safety-aligned testbed with established attack and evaluation tooling.
AdvBench \citep{zou2023universal} is split into 420 harmful instructions
for attribution, AdvBench$_{\text{attr}}$, and 100 for evaluation,
AdvBench$_{\text{eval}}$. For safety attribution, we collect refusal
responses from the aligned model on AdvBench$_{\text{attr}}$. Utility
attribution uses Alpaca-Cleaned, a cleaned version of Stanford Alpaca
\citep{taori2023alpaca}, with safety-related examples removed.

\paragraph{Safety evaluation.}
We evaluate attack success rate (ASR) on
$\mathrm{AdvBench}_{\mathrm{eval}}$. Following
\citet{zou2023universal}, an attack is considered successful when the
generated response lacks key refusal patterns. We report \emph{Basic ASR}
on harmful prompts without an adversarial suffix and \emph{GCG ASR} using
GCG adversarial suffixes~\citep{zou2023universal}.

\paragraph{Utility evaluation.}
General utility is measured using
tinyBenchmarks~\citep{maiapolo2024tinybenchmarks}. The unmodified model
achieves 52.2\% accuracy, and we report post-intervention accuracy to
measure broader capability degradation.

\paragraph{Interventions.}
For low-rank localization, we vary the retained fraction of the
safety-associated weight intervention and its architectural scope. For
parameter-level localization, safety- and utility-important weights are
combined using the set-difference criterion in
Eq.~\ref{eq:set_difference}. We evaluate interventions across all
projection modules, MLP projections, and selected individual projection
types. Reported weight fractions are measured relative to the full model.

\section{Results}
\label{sec:results}
\subsection{Low-Rank Intervention Sparsity and Localization}
Table~\ref{tab:lowrank_results} shows that retaining only the top 10\%
of the low-rank intervention preserves substantial safety degradation,
while effectiveness drops sharply at 1\%, indicating concentration among
higher-magnitude modifications. Restricting the intervention to MLP
projections retains much of the effect, and \texttt{down\_proj} alone
remains sufficient to induce notable safety degradation. Utility remains
close to the 52.2\% baseline across the more localized settings.
\begin{table}[ht]
\caption{Low-rank intervention results across sparsity levels and
architectural components. Utility denotes tinyBenchmarks accuracy;
the unmodified model achieves 52.2\%.}
\label{tab:lowrank_results}
\centering
\small
\begin{tabular}{lcccc}
\toprule
\textbf{Modules} &
\textbf{Top fraction} &
\textbf{Basic ASR} &
\textbf{GCG ASR} &
\textbf{Utility} \\
\midrule
All modules             & 1.0   & 87\% & 98\% & 49.3\% \\
All modules             & 0.1   & 71\% & 85\% & 50.1\% \\
All modules             & 0.01  & 36\% & 25\% & 50.7\% \\
All modules             & 0.001 & 20\% & 18\% & 51.8\% \\
\midrule
MLP only                & 1.0   & 77\% & 89\% & 49.8\% \\
MLP only                & 0.1   & 54\% & 64\% & 51.1\% \\
\texttt{down\_proj}     & 1.0   & 51\% & 43\% & 51.2\% \\
\texttt{down\_proj}     & 0.1   & 32\% & 26\% & 51.4\% \\
\bottomrule
\end{tabular}
\end{table}
\subsection{Parameter-Level Safety--Utility Localization}
\begin{table}[t]
\caption{Sparse interventions using parameter-level safety--utility
localization. Weight fraction is measured relative to the full model.
Utility denotes tinyBenchmarks accuracy; the unmodified model achieves
52.2\%.}
\label{tab:param_results}
\centering
\small
\begin{tabular}{lcccc}
\toprule
\textbf{Target component} &
\textbf{Weights (\%)} &
\textbf{Basic ASR} &
\textbf{GCG ASR} &
\textbf{Utility} \\
\midrule
All modules
& 2.95 & 91\% & 97\% & 47.1\% \\
MLP only
& 1.87 & 86\% & 94\% & 49.7\% \\
\texttt{down\_proj} + \texttt{o\_proj}
& 0.23 & 58\% & 60\% & 51.3\% \\
\texttt{down\_proj}
& 0.19 & 53\% & 56\% & 51.6\% \\
\texttt{o\_proj}
& 0.04 & 20\% & 14\% & 52.0\% \\
\bottomrule
\end{tabular}
\end{table}
Table~\ref{tab:param_results} shows strong architectural concentration in
the parameter-level localization results. Restricting the intervention from
all modules to MLP projections retains most of the safety effect, while
further localization to \texttt{down\_proj} and \texttt{o\_proj} reduces
the intervention to 0.23\% of model weights while still achieving 58\%
Basic ASR and 60\% GCG ASR. In particular, \texttt{down\_proj} alone
modifies only 0.19\% of the model while yielding 53\% Basic ASR and 56\%
GCG ASR, whereas \texttt{o\_proj} alone is substantially weaker. Across
both localization approaches, \texttt{down\_proj} consistently emerges as
the most prominent safety-sensitive component, with utility remaining near
the unmodified baseline for these highly localized interventions.

\section{Discussion}
Our results indicate that safety-sensitive parameters are highly non-uniform
across LLaMA-2-7B-Chat. Both localization approaches consistently identify
the MLP \texttt{down\_proj} as the most prominent component, with
\texttt{o\_proj} providing a smaller complementary contribution. The
parameter-level intervention targeting only \texttt{down\_proj} modifies
0.19\% of model weights while still producing substantial safety
degradation, with tinyBenchmarks accuracy remaining close to the
unmodified baseline. This suggests that the safety-relevant parameter
space is substantially smaller than the full model.

Parameter sparsity, however, should not be interpreted directly as
bit-level vulnerability. Even 0.19\% of a 7B-parameter model corresponds
to millions of weights, whereas a practical hardware fault attack would
ideally require only a small number of physically realizable bit
modifications. Bit-level effects will also depend on numerical
representation, memory architecture, and fault mechanism. Our current
experiments therefore localize a reduced safety-sensitive parameter space,
but do not yet demonstrate an end-to-end hardware attack.

The utility results provide initial evidence that localized interventions
can degrade safety without proportionally degrading general capability,
although tinyBenchmarks provides only a compact estimate of utility.
These findings are particularly relevant to on-device and
resource-constrained deployments, including language models used as
components of agentic systems, where locally stored parameters may remain
exposed throughout repeated model interactions. Future work will study
whether these localized regions are enriched for high-impact bit-level
faults, compare against random and magnitude-based baselines, and examine
selective integrity protection for safety-sensitive regions in such
deployments.

\section{Conclusion}
We show that safety-sensitive parameters in LLaMA-2-7B-Chat are structurally concentrated, with \texttt{down\_proj} consistently emerging as the most prominent component across both low-rank and parameter-level localization. Parameter-level localization reduces the intervention to only 0.19\% of model weights while achieving 53\% Basic ASR and 56\% GCG ASR, with utility remaining close to the unmodified baseline. These results suggest that safety-relevant behavior depends disproportionately on a small and structured portion of the parameter space, creating a reduced candidate surface for subsequent fault analysis. This concentration is particularly relevant for on-device and resource-constrained language models, including models used as components of agentic systems, where locally stored parameters may be exposed to hardware reliability and integrity faults. Future work will investigate whether these localized regions contain a similarly sparse set of high-impact, physically realizable bit-level vulnerabilities and whether selective integrity protection can efficiently harden such safety-sensitive regions.

\clearpage
\bibliographystyle{plainnat}
\bibliography{references}

\end{document}